%% file: output.tex
\documentclass[lettersize,journal]{IEEEtran}
\usepackage[utf8]{inputenc}
\usepackage{epsfig}
\usepackage{graphicx}
\usepackage{amsmath}
\usepackage{amssymb}
\usepackage{color}
\usepackage{multirow}
\usepackage{algorithmic}
\usepackage{algorithm}
\usepackage{bbding} 
\usepackage{array}
\usepackage{textcomp}
\usepackage{stfloats}
\usepackage{url}
\usepackage{verbatim}
\usepackage{graphicx}
\usepackage{cite}
\usepackage{tablefootnote}
\usepackage{ulem} 
\usepackage{mathrsfs}
\usepackage{ulem}

\usepackage{hyperref}

\begin{document}

%%%%%%%%% TITLE
\title{Dynamic Clustering Transformer Network for Point Cloud Segmentation}

\author{
Dening Lu,
Jun Zhou,
Kyle (Yilin) Gao,
Dilong Li,
Jing Du,
Linlin Xu,
Jonathan Li

\thanks{\textit{(Corresponding authors: Linlin Xu; Jonathan Li.)}}

\thanks{Dening Lu is with the Department of Systems Design Engineering, University of Waterloo, Waterloo, Ontraio N2L 3G1, Canada (e-mail: d62lu@uwaterloo.ca).}
}

\maketitle
%%%%%%%%% ABSTRACT
\begin{abstract}
   Point cloud segmentation is one of the most important tasks in computer vision with widespread scientific, industrial, and commercial applications. The research thereof has resulted in many breakthroughs in 3D object and scene understanding. Previous methods typically utilized hierarchical architectures for feature representation. However, the commonly used sampling and grouping methods in hierarchical networks are only based on point-wise three-dimensional coordinates, ignoring local semantic homogeneity of point clusters. Additionally, the prevalent Farthest Point Sampling (FPS) method is often a computational bottleneck. To address these issues, we propose a novel 3D point cloud representation network, called Dynamic Clustering Transformer Network (DCTNet). It has an encoder-decoder architecture, allowing for both local and global feature learning. Specifically, we propose novel semantic feature-based dynamic sampling and clustering methods in the encoder, which enables the model to be aware of local semantic homogeneity for local feature aggregation.
   % the encoder of DCTNet has three main stages. Each stage consists of a dynamic clustering-based Local Feature Aggregating (LFA) block and a Transformer-based Global Feature Learning (GFL) block. In the LFA block, we propose novel semantic feature-based dynamic sampling and clustering algorithms for local feature extraction. 
   Furthermore, in the decoder, we propose an efficient semantic feature-guided upsampling method. Our method was evaluated on an object-based dataset (ShapeNet), an urban navigation dataset (Toronto-3D), and a multispectral LiDAR dataset, verifying the performance of DCTNet across a wide variety of practical engineering applications. The inference speed of DCTNet is 3.8-16.8$\times$ faster than existing State-of-the-Art (SOTA) models on the ShapeNet dataset, while achieving an instance-wise mIoU of $86.6\%$, the current top score. Our method similarly outperforms previous methods on the other datasets,
   %. Extensive experiments demonstrate the effectiveness of our DCTNet, 
   verifying it as the new State-of-the-Art in point cloud segmentation.
\end{abstract}


%%%%%%%%% BODY TEXT
\input{Intro}

\input{Related_work}

\input{method}
\input{result_discuss}
\input{conclusion}

\bibliographystyle{IEEEtran}
\bibliography{egbib}

\vfill
\end{document}

%% file: Intro.tex
\section{Introduction}
\label{sec:introduction}

Semantic segmentation of 3D point clouds has been widely used in various applications, such as autonomous driving, robotics, city information modeling, engineering survey and mapping. Compared to images formed by 2D pixels, 3D point clouds are more complicated and more flexible. The design of advanced methods tailored to the characteristics of 3D point clouds for enhanced 3D point cloud processing is currently one of the most active research topics in computer vision.

\begin{figure}[t]
  \centering
  \includegraphics[width=\linewidth]{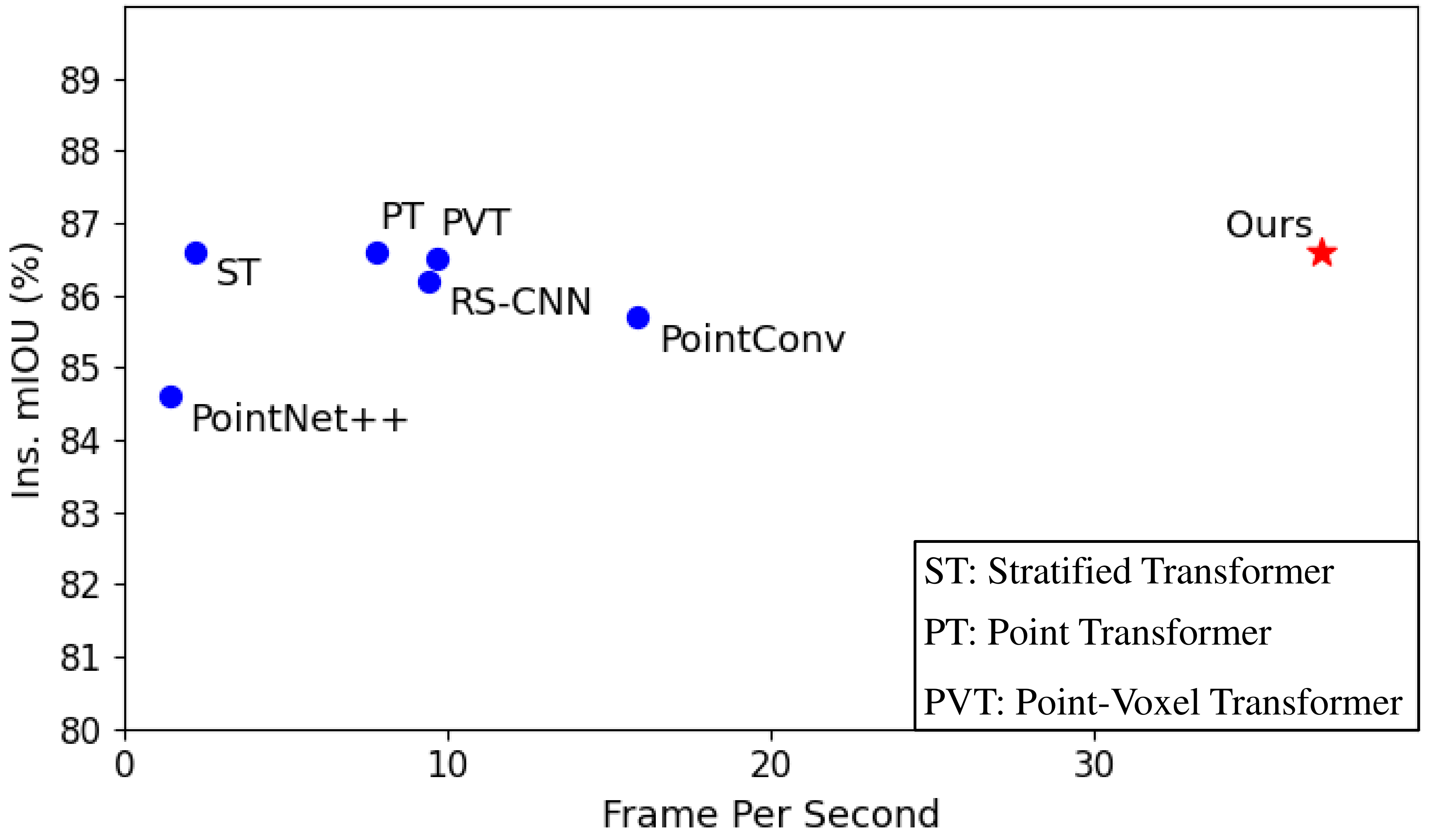}
  \caption{Performance comparison of SOTA methods on the ShapeNet dataset.
  \label{fig:eff}}
\end{figure}

Existing methods for 3D point cloud segmentation can be generally divided into three categories:  view-based \cite{kundu2020virtual, robert2022learning, mascaro2021diffuser, antonello2018multi, dai20183dmv}, voxel-based \cite{maturana2015voxnet, riegler2017octnet, zhou2018voxelnet, zhang2022pvt}, and point-based \cite{qi2017pointnet, qi2017pointnet++, wang2019dynamic, thomas2019kpconv, zhao2021point, guo2021pct, lai2022stratified}.
Most of them utilized hierarchical structures for point cloud processing, focusing on local feature extraction but often ignoring long-range context dependency modeling \cite{lai2022stratified}. The hierarchical structure typically involves two key steps: point cloud sampling and grouping. Currently, most hierarchical point cloud processing methods use the Farthest Point Sampling (FPS) \cite{qi2017pointnet++} algorithm, sampling points evenly across the geometric space.
However, FPS only focuses on the geometric properties of point clouds, ignoring their semantic features. This causes neural networks to de-emphasize some fine-level object parts with significant semantic information. Moreover, FPS is very time-consuming, often causing a computational bottleneck. Additionally, after downsampling, $k$-Nearest Neighborhood ($k$NN) and ball query are widely used for the point cloud grouping. However, such grouping methods are still strictly based on the geometric properties of points. In this situation, the local feature aggregation tends to be disturbed by semantic heterogeneity in local neighborhoods, especially for points at the boundaries of adjacent parts. Similar to superpixel in image processing, SuperPoint Graph (SPG) proposed in \cite{landrieu2018large} was design for clustering semantically homogeneous points into the same group. It was able to describe in detail the relationship between adjacent objects. However, 
% SPG is only performed as a pre-processing step before the deep learning network \cite{sun2022superpoint}, with initial geometric properties such as 3D position, color, intensity, etc. 
as a pre-processing step before the deep learning network \cite{sun2022superpoint}, SPG fails to perform dynamic sampling and clustering for hierarchically extracted semantic features at different stages in the network, limiting its performance.

To address the aforementioned issues, we propose a novel hierarchical point cloud representation framework for 3D semantic segmentation. It combines both dynamic clustering-based Local Feature Aggregating (LFA) blocks and Transformer-based Global Feature Learning (GFL) blocks. We introduce novel semantic feature-based dynamic sampling and clustering methods to LFA blocks, focusing on local semantic homogeneity of point clusters belonging to any particular object and improving algorithm efficiency. For GFL blocks, we use the dual-attention Transformer to capture long-range context dependencies. 
%##########################################################

The main contributions of our work are as follows:
\begin{itemize}
    \item We propose the novel Semantic feature-based Dynamic Sampling (SDS) and Clustering (SDC) methods for dynamic token generation and local feature aggregating. The proposed approaches can not only better identify local semantic homogeneity of 3D objects for improved semantic segmentation, but also can greatly improve the computational efficiency compared to traditional sampling and grouping approaches. 
    \item We design a Transformer-based hierarchical 3D representation framework (named DCTNet) for point cloud segmentation. The encoder-decoder architecture is highly efficient to capture local-global information due to its dynamic token generation mechanism in LFA blocks and Transformer-based GFL blocks. In the decoder, we propose an efficient semantic feature-guided upsampling method, ensuring simple yet highly accurate upsampling operation. 
    \item We conducted expensive ablation studies and comparative studies on different public segmentation datasets with various previous SOTA approaches. The ablation results prove the effectiveness of each building block of DCTNet, as well as the neural architecture as a whole. The benchmark results indicate our DCTNet can be considered the new State-of-the-Art in 3D point cloud segmentation.  
\end{itemize}

%% file: Related_work.tex
\begin{figure*}[htbp]
  \centering
  \includegraphics[width=0.95\linewidth]{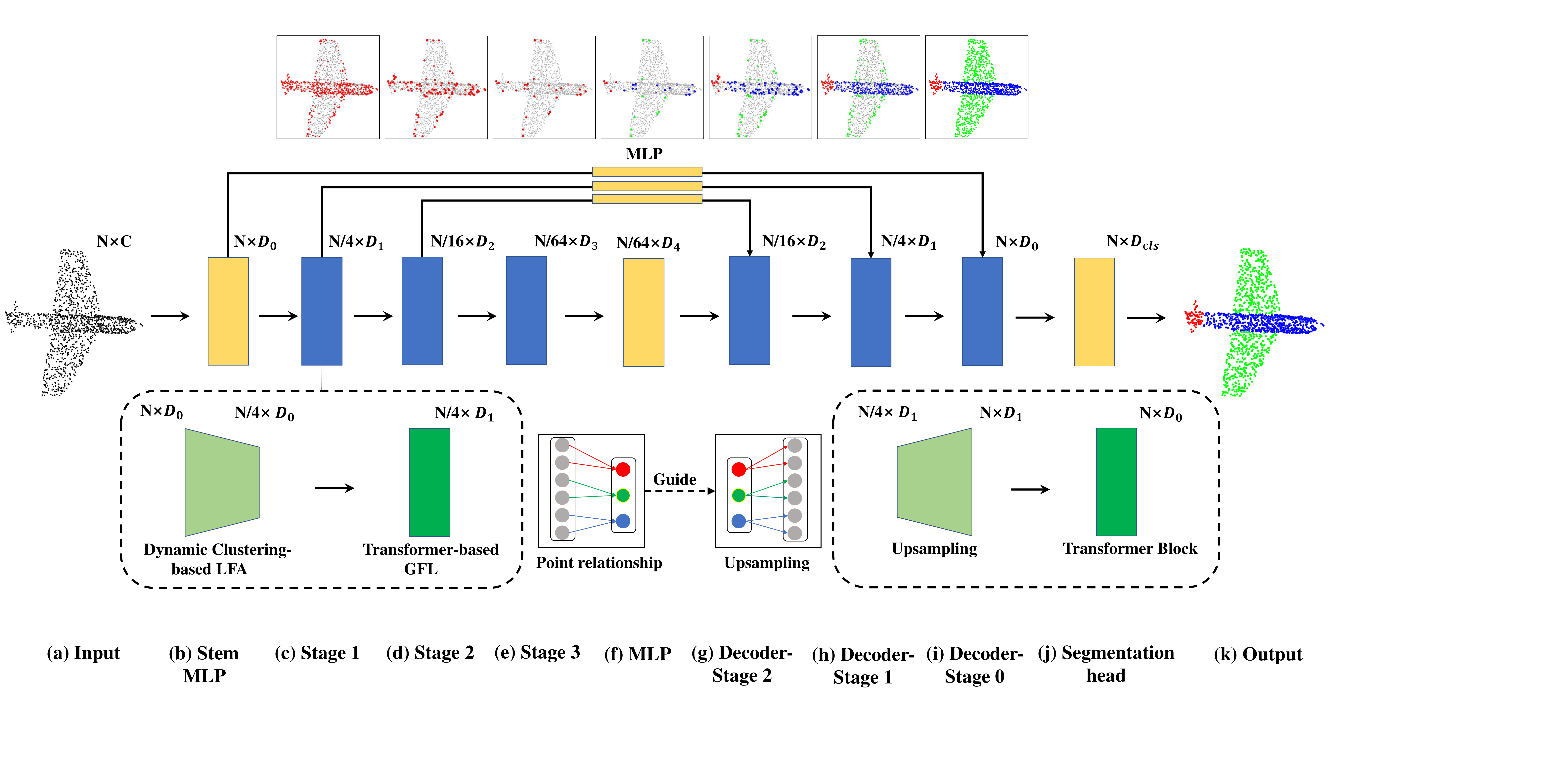}
  \caption{Hierarchical encoder-decoder structure of DCTNet for point cloud segmentation. Dynamic clustering-based LFA blocks and Transformer-based GFL blocks are designed for local-global feature representation. For the top seven subfigures, the first three show the hierarchically sampled points in the encoder, and the last four show the semantic information of upsampling points in the decoder. 
  % XXXXXXXXXXXXXXX Dening - It would be nice if you can explain a lit this figure to ensure that the readers can understand the architecture and realize the novelty of this architecture without reading the text. XXXXXXXXXXXXXXXXXXXXXXXXXXXXXXXXXXX
  \label{fig:overview}}
\end{figure*}

\section{Related Work}
\label{sec:relatedwork}

\subsection{Point Cloud Segmentation}
View-based segmentation methods \cite{kundu2020virtual, robert2022learning, mascaro2021diffuser, antonello2018multi, dai20183dmv} 
% were motivated by the great success of view-based point cloud classification. It
had three main steps: multi-view projection, image feature extraction, and multi-view unprojection \cite{hamdi2021voint}. However, view-based methods tended to incur geometric information loss during projection, which hindered performance improvement. Similar to image processing, voxel-based methods such as VoxNet \cite{maturana2015voxnet} and its variants \cite{riegler2017octnet, zhou2018voxelnet} proposed to use the regular volumetric grid to represent unstructured point clouds. 
% As such, unorder point clouds were transformed into regular voxels, which could be processed directly by 3D convolutions. 
However, such methods tended to incur high computational costs, as well as result in geometric information loss. PointNet \cite{qi2017pointnet}, PointNet++ \cite{qi2017pointnet++}, and their variants \cite{wang2019dynamic, thomas2019kpconv, landrieu2018large} designed point-based methods to process raw point clouds directly. The basic idea was to use a series of shared Multi-Layer Perceptrons (MLPs), and max/mean pooling operations for feature extraction and aggregation. Despite achieving great success in point cloud processing, most point-based methods extracted only point-wise or local features because of the inductive bias of locality, making it challenging for these methods to learn global information. Recently, many 3D Transformers tailored to point cloud segmentation were proposed. They are also point-based methods. Point Transformer (PT) \cite{zhao2021point} and Point Cloud Transformer (PCT) \cite{guo2021pct} designed pure Transformer architectures for feature learning and aggregation. 
% Dual Transformer Network (DT-Net) proposed to combine both point- and channel-wise self-attention mechanisms for stronger global feature enhancement. 
Stratified Transformer \cite{lai2022stratified} incorporated the ideas from Swin Transformer \cite{liu2021swin} in image processing to point cloud segmentation, achieving satisfactory results. 

\subsection{Dynamic Clustering for Feature Aggregation}
Different from commonly-used grouping methods like $k$NN, dynamic clustering focuses on the similarity of neighboring points at the feature level. DGCNN \cite{wang2019dynamic} proposed to achieve dynamic point clustering in each layer of the network based on dynamically updated EdgeConv. Similarly, DPFA-Net \cite{chen2022background} and DSACNN \cite{song2022dsacnn} utilized feature-level $k$NN to generate dynamic neighborhoods. GAC \cite{wang2019graph} proposed to cluster the most relevant neighboring points by dynamically assigning proper attentional weights to them. However, for hierarchical processing algorithms like GAC \cite{wang2019graph}, their clustering centers are often obtained by geometry-level downsampling methods like FPS, instead of feature level. This limits the performance of dynamic clustering to some extent. TCFormer \cite{zeng2022not} in image processing introduced feature-level cluster center selecting and dynamic token clustering methods to the vision Transformer. It achieved satisfactory results in various human-centric tasks. Inspired by TCFormer \cite{zeng2022not}, our work proposes novel and efficient dynamic sampling and clustering methods for local feature aggregation, combined with the cross-attention Transformer. Further, we utilize the dynamic clustering relationship in the encoder to guide the upsampling process in the decoder.

%% file: method.tex
% ##############################################################
\begin{figure*}[htbp]
  \centering
  \includegraphics[width=0.9\linewidth]{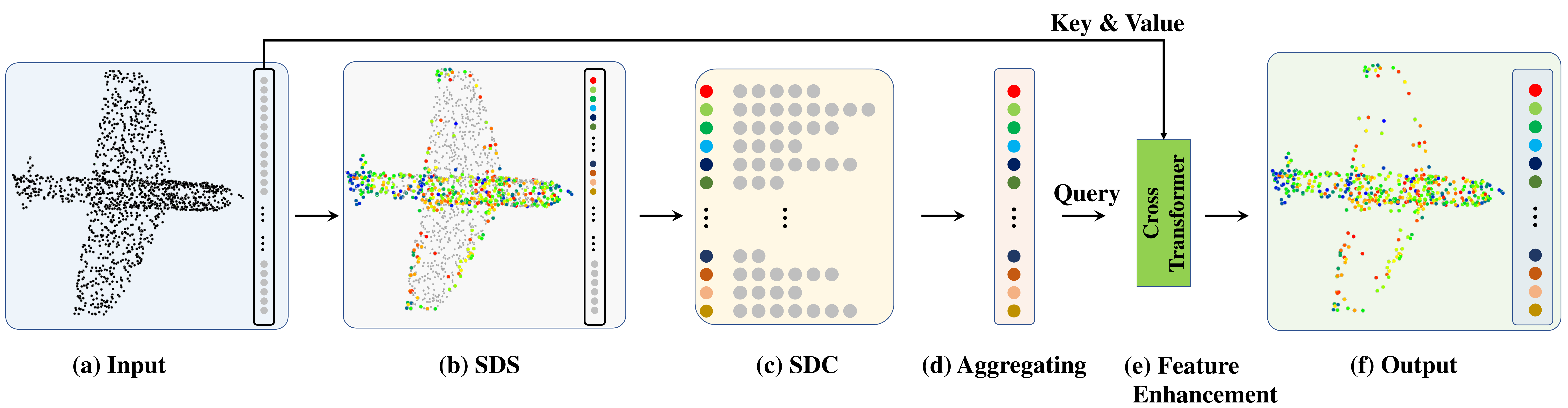}
  \caption{Pipeline of the dynamic clustering-based LFA block. Semantic feature-based Dynamic Sampling (SDS) and Clustering (SDC) are proposed to ensure local semantic homogeneity in clusters, facilitating local feature aggregating.
  \label{fig:LFA}}
\end{figure*}
% ##############################################################

\section{Dynamic Clustering Transformer Network}
\label{sec:method}
This section shows the details of DCTNet. We first present the overall pipeline, then introduce its main blocks: dynamic clustering-based LFA block, Transformer-based GFL block, and semantic feature-guided upsampling block.

\subsection{Overview}
The overall pipeline of DCTNet is shown in Fig. \ref{fig:overview}.
The original point cloud with/without normal is taken as input to the encoder. Firstly, a stem MLP block \cite{qian2022pointnext} is designed to project the input data into a higher-dimension space. 
Secondly, the projected features are fed into several stages in a hierarchical manner for local and global feature extraction. Each stage in the encoder consists of two blocks: a dynamic clustering-based LFA block and a Transformer-based GFL block. 
Thirdly, the extracted features by the aforementioned stages are taken as input to the decoder. 
Specifically, the decoder follows the U-Net design, symmetric to the encoder structure described above. As shown in Fig. \ref{fig:overview}, each stage in the decoder consists of two blocks: a semantic feature-guided upsampling block and a Transformer-based GFL block which is exactly the same as the corresponding block in the encoder. 
Lastly, an MLP head layer is used to get the final prediction for each point, which consists of two linear layers with batch normalization and ReLU.
% Since the local information extracted in the encoder is introduced to the decoder modules by the skip connection, there is no additional LFA block in the decoder. 
% We note that the number of modules and the sampling rates can vary according to the specific applications, for example, to construct light-weight backbones for fast processing.

\subsection{Dynamic Clustering-based LFA Block}
\label{subsec:LFA}
The dynamic clustering-based LFA block is designed to achieve discriminative local feature extraction. Inspired by TCFormer \cite{zeng2022not}, our LFA block consists of three key steps: point cloud dynamic clustering, local feature aggregating, and feature enhancement. The pipeline of the LFA block is shown in Fig. \ref{fig:LFA}. The first step is to achieve point cloud sampling and generate semantically homogeneous clusters for sampling points. The second step is to aggregate the point features in the same cluster. The last step is to establish the connection between the aggregated sampling points and input features, enhancing the sampling point features and mitigating feature loss caused by aggregating.

\textbf{Point Cloud Dynamic Clustering.}
We propose SDS and SDC methods for point cloud sampling and clustering. For our implementation of SDS, given an input point set $P = \left \{ p_{i} \right \}_{i=1}^{N} \in R^{N \times D}$, where $D$ is the dimension of the input feature, we first compute the local density $d_{i}$ of each point $p_{i}$ according to its $k$-nearest neighborhood $\Phi_{i}$ in the feature space:
\begin{equation}
d_{i}= exp \left ( -\frac{1}{k}\sum_{p_{j} \in \Phi_{i}} \left \| p_{i}-p_{j} \right \| ^{2} \right ).
\end{equation}
According to $P = \left \{ p_{i} \right \} _{i=1}^{N}$, we denote $\Gamma = \left \{d_{i} \right \} _{i=1}^{N}$.

\begin{figure}[htbp]
  \centering
  \includegraphics[width=\linewidth]{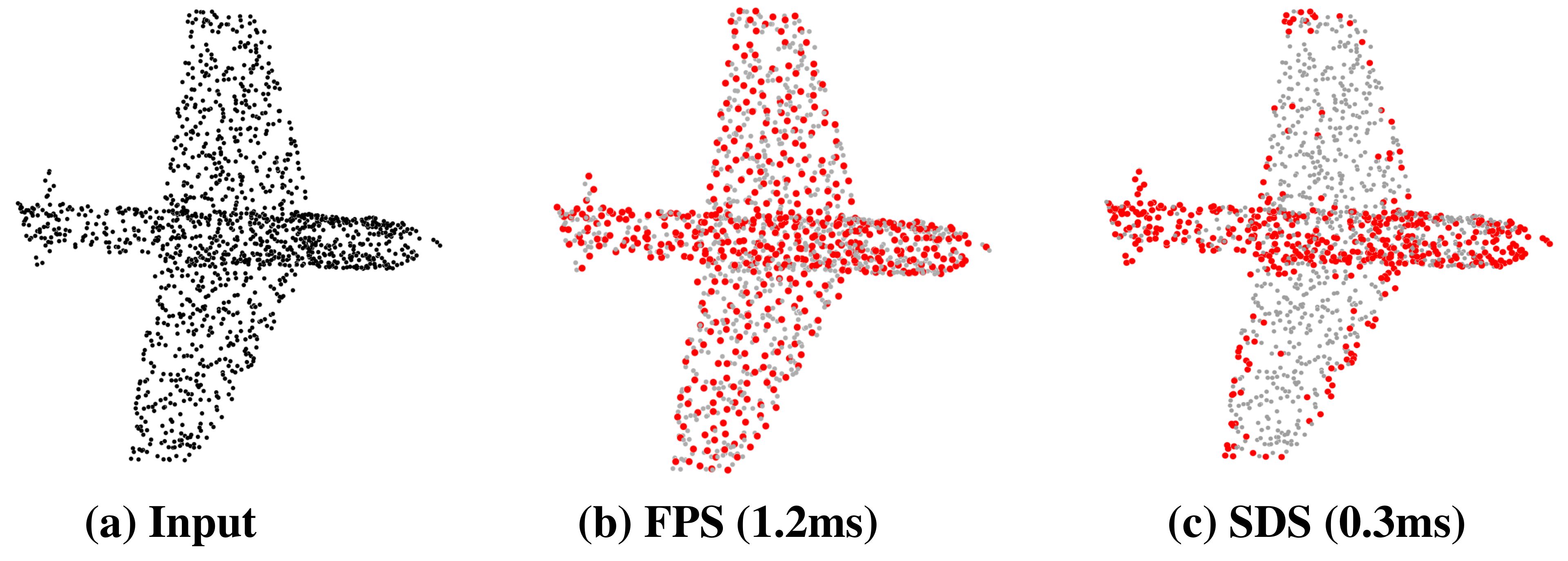}
  \caption{Comparison of sampling results. Compared with FPS, our SDS focuses on key discriminative geometric areas, retaining fewer points in the flat areas but more points in the nose, tail, and wing contours. Additionally, it is more efficient, about 4$\times$ faster than FPS.
  \label{fig:sampling}}
\end{figure}

Secondly, we calculate a distance indicator $\delta_{i}$ for $p_{i}$ \cite{du2016study}, which can be expressed as:
\begin{equation}
\delta_{i}= \left\{\begin{matrix}
  min_{j:d_{j} \in \Omega_{i}}\left \| p_{i}-p_{j} \right \|^{2}, & if\;\Omega_{i} \neq \emptyset   \\
  max_{j:d_{j} \in \Gamma }\left \| p_{i}-p_{j} \right \|^{2},  & otherwise
\end{matrix}\right.
\end{equation}
where $\Omega_{i} = \left \{ d_{j}\in \Gamma |\forall 
 d_{j}>d_{i} \right \}$. According to this equation, $\delta_{i}$ can be understood as the minimal feature distance between $p_{i}$ and any other points with higher local density. For the point with the highest local density, its distance indicator is defined as the maximal feature distance between it and any other points.
Given $d_{i}$ and $\delta_{i}$, we combine them to get the score of each point, which can be expressed as $\delta_{i} \times d_{i}$. A higher score means this point has a more representative feature and then is more suitable to be selected as the sampling point. Therefore, according to the sampling rate, we choose the points with the highest scores as sampling points. Based on semantic features, the sampling points are dynamically updated in each stage of the network.
As shown in Fig. \ref{fig:sampling}, compared with FPS, our SDS retains fewer points in the flat areas but more points in the key areas, such as the nose, tail, and contours of the wings. This could provide more useful information for network learning. Additionally, it is more efficient, about 4$\times$ faster than FPS. 

\begin{figure}[htbp]
  \centering
  \includegraphics[width=\linewidth]{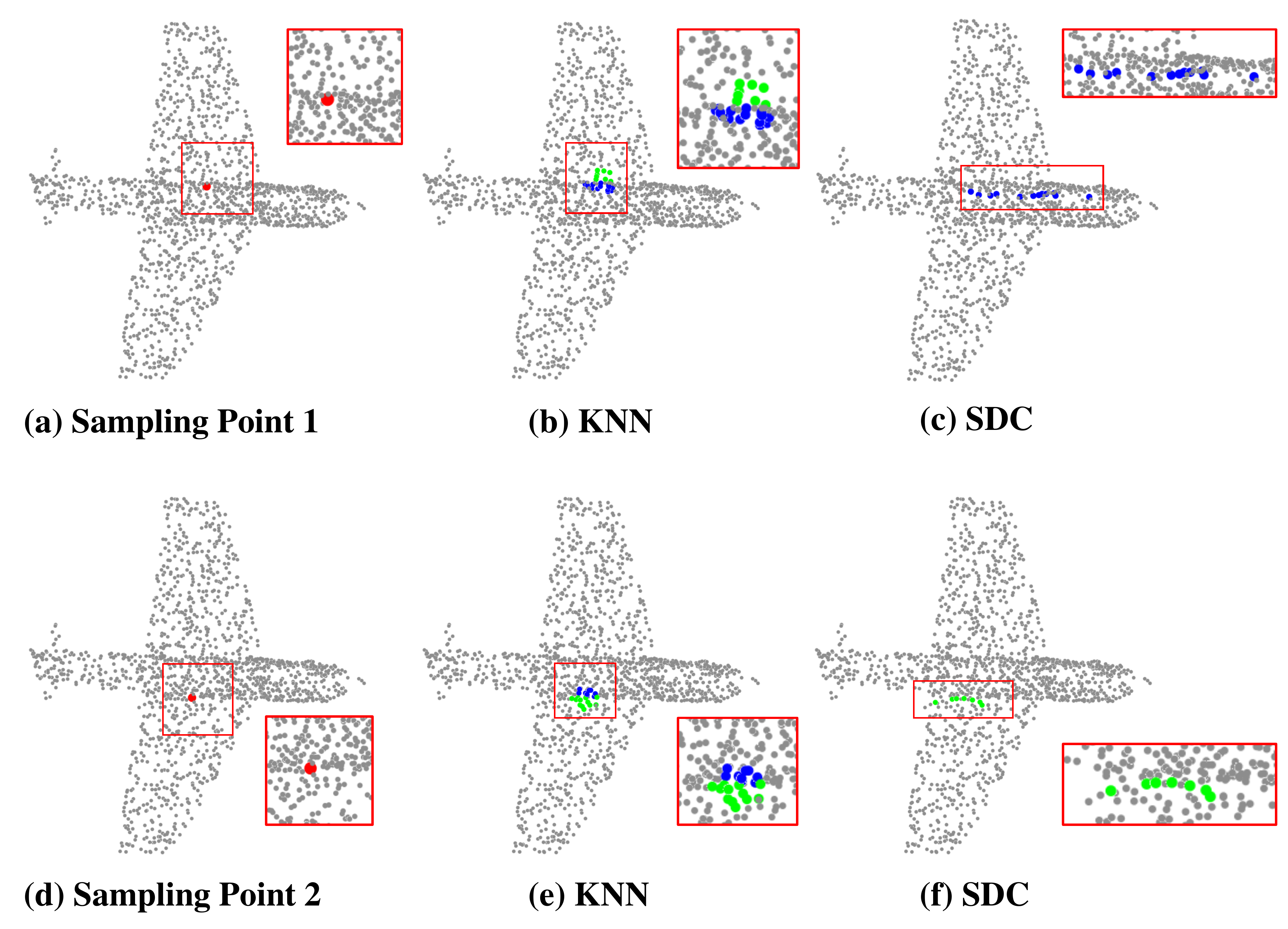}
  \caption{Grouping results of different methods. Compared with $k$NN, our SDC is able to cluster points with similar semantic information, ensuring local semantic homogeneity within the same group. The fuselage and wings are colored blue and green respectively.
  \label{fig:clustering}}
\end{figure}

Thirdly, given the sampling point set $\mathbb{S} = \left \{ s_{i} \right \} _{i=1}^{S} \in R^{S \times D}$, we design the SDC method to construct a cluster for each $s_{i}$. Specifically, in the feature space, small feature distances mean similar semantic information. Therefore, we assign every point in $P$ to the nearest sampling point in $\mathbb{S}$ based on the feature distances. As such, each $s_{i}$ has a cluster $C_{si}$ with local semantic homogeneity, facilitating local feature aggregating. According to dynamically generated sampling points, the clustering process is also dynamically updated in each stage based on semantic features. As shown in Fig. \ref{fig:clustering}, for points at the boundaries of fuselage and wings, $k$NN grouping tends to include points from two different parts into the same group, which may disturb the local feature aggregating. However, our SDC method is able to cluster points with similar semantic information, ensuring local semantic homogeneity within the same group.

\textbf{Local Feature Aggregating.}
Given clusters $\mathbb{C} = \left \{ C_{si} \right \} _{i=1}^{S} $, we utilize a learning-based weighted average algorithm to achieve the local feature aggregating. 

Since the cluster points in $C_{si}$ have similar semantic features, an intuitive method is to average them directly, obtaining the $s_{i}$ with local information, which can be can be expressed as:
\begin{equation}
s_{i}= \operatorname*{\textit{average}}\limits_{j \in C_{si}} \, ( C_{si\_{j}}),
\end{equation}
where $C_{si\_{j}}$ denotes the $j$-th cluster points in $C_{si}$. However, it is still hard for points in the same cluster to have the same importance for the network. This simple average may lead to information loss. Therefore, we implement a learnable attention score set $A = \left \{ a_{i} \right \} _{i=1}^{N} $ \cite{zeng2022not} for all points in $P$. Then for the cluster $C_{si}$, $s_{i}$ can be expressed as:
\begin{equation}
s_{i}= \frac{\sum_{j \in C_{si}}exp(a_{j})C_{si\_{j}}}{ \sum_{j \in C_{si}}exp(a_{j})},
\end{equation}
where $a_{j}$ is the learnable attention score of $C_{si\_{j}}$. As such, $s_{i}$ is able to describe the local semantic information more accurately. As such, the aggregated sampling point set $\mathbb{S} =  \left \{ s_{i} \right \} _{i=1}^{S}$ is obtained. The relationship between the sampling points and cluster points is also stored for the point cloud upsampling in the decoder.

\textbf{Feature Enhancement.}
Given the sampling point set $\mathbb{S}$, we design a cross-attention Transformer to establish the connection between $\mathbb{S}$ and input features $P$, enhancing sampling point features and mitigating information loss caused by the aggregating process.

Specifically, as shown in Fig. \ref{fig:LFA}, we first generate $Query$ matrix based on $\mathbb{S}$, and $Key, Value$ matrices based on $P$: 
\begin{equation}
\begin{aligned}
Q =  \mathbb{S} W_{Q},\\
K =  P W_{K},\\
V =  P W_{V},\\
\end{aligned}
\end{equation}
where $Q, K, V$ denote $Query, Key$, and $Value$ matrices. $W_{Q} W_{K}, W_{V}$ are learnable weight matrices. Then, the attention map $M$ can be formulated as:
\begin{equation}
M = softmax(\frac{QK^{T}}{\sqrt{D}}+A).
\end{equation}
The size of $QK^{T}$ is $S \times N$, where each element $m_{i, j}$ represents the feature similarity between $i$-th sampling point in $\mathbb{S}$ and $j$-th input point in $P$. However, as shown above, the size of $A$ is $1 \times N$, which is inconsistent with $QK^{T}$. Therefore, for our implementation, we repeat $A$ along rows, extending its size to $S \times N$. The element addition between $QK^{T}$ and $A$ means that both feature similarity and point importance are considered in our cross-attention Transformer. Finally, the enhanced sampling point set $\mathbb{S}$ can be generated by multiplying $M$ and $V$, with the size of $S \times D$.

\subsection{Transformer-based GFL Block}
\label{subsec:GFL}
We use the Transformer to achieve global feature learning, thanks to its remarkable ability of long-range context dependency modeling. The dual-attention Transformer \cite{han2022dual, lu20223dpct} has been proven more effective in global feature learning than vanilla point-wise or channel-wise Transformers. Therefore, we use the dual-attention Transformer in our GFL block. The Point-wise Self-Attention (PSA) in the dual-attention Transformer is used to build the spatial relationship between points, achieving long-range context dependency modeling. Similarly, the Channel-wise Self-Attention (CSA) in the dual-attention Transformer is used to explore the difference between feature channels, highlighting the role of interaction across various channels \cite{han2022dual}. By combing these two kinds of self-attention mechanisms, our GFL block is able to capture global features from multiple perspectives.

PSA and CSA have similar algorithm flows. Specifically, taking the sampling point set $\mathbb{S}$ as input, we first project it into three different feature spaces to generate $Query$, $Key$, and $Value$ matrices:
\begin{equation}
\begin{aligned}
Query =  \mathbb{S} W_{Q},\\
Key =  \mathbb{S} W_{K} ,\\
Value =  \mathbb{S} W_{V} ,\\
\end{aligned}
\end{equation}
where $W_{Q}, W_{K}, W_{V}$ are learnable weight matrices. Secondly, for the PSA, the attention map $M_{P}\in R^{S \times S}$ can be formulated as:
\begin{equation}
M_{P} = softmax(\frac{QK^{T}}{\sqrt{D}}+B),
\end{equation}
where $Q,K$ denote the $Query$, $Key$ matrices, and $B$ is a learnable position encoding matrix defined by \cite{zhao2021point}. 
$M_{P}$ and $Value$ matrices are multiplied to generate the new feature map $F_{P}$ as the output of PSA, of the same size as $\mathbb{S}$.
Thirdly, for the CSA, the attention map $M_{C}\in R^{D \times D}$ can be formulated as:
\begin{equation}
M_{C} = softmax(\frac{K^{T}Q}{\sqrt{D}}).
\end{equation}
$Value$ and $M_{C}$ matrices are multiplied to generate the new feature map $F_{C}$ as the output of CSA, of the same size as $\mathbb{S}$.

Given global feature maps $F_{P}$ and $F_{C}$, we combine them by the element-wise addition:
\begin{equation}
F_{G} = F_{P} + F_{C}.
\end{equation}
Lastly, we apply a skip connection between $F_{G}$ and the input feature set $\mathbb{S}$:
\begin{equation}
F_{G} = \mathbb{S} + LBR(F_{G}),
\end{equation}
where $F_{G}$ is the final global feature map, and $LBR$ denotes the combination of $Linear$, $BatchNorm$, and $ReLU$.

\begin{figure*}[htbp]
  \centering
  \includegraphics[width=\linewidth]{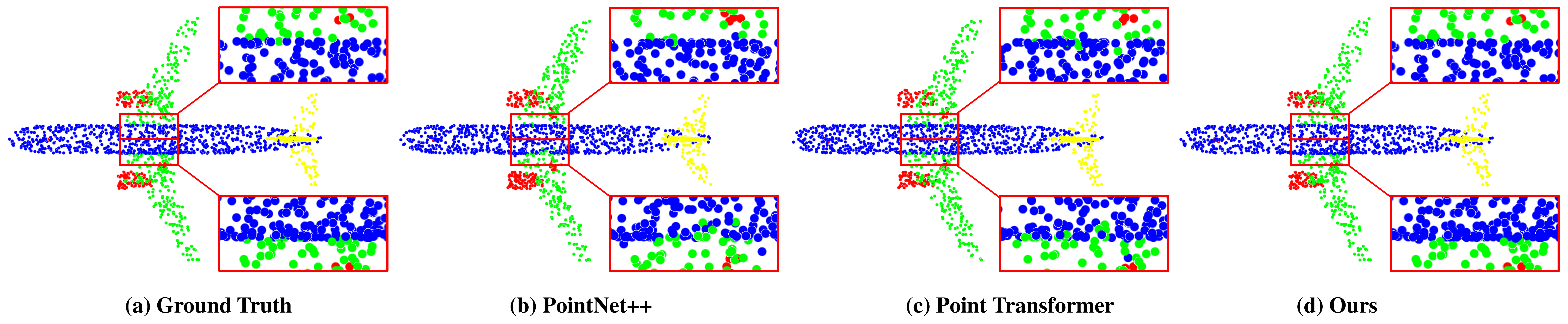}
  \caption{Airplane segmentation results from different methods. Our method achieves the best results at the boundaries of adjacent parts. The fuselage, wings, engines, and tail are colored blue, green, red, and yellow respectively.
  \label{fig:comparison}}
\end{figure*}

\subsection{Semantic Feature-guided Upsampling Block}
\label{subsec:upsampling}
As shown in Fig. \ref{fig:overview}, since the relationship between sampling points and cluster points has been stored in the encoder, point cloud upsampling can be easily achieved by assigning the semantic features of sampling points to corresponding cluster points. Since the relationship is obtained by semantic feature-based clustering, the upsampling process is named semantic feature-guided upsampling.

As such, compared with commonly used point cloud interpolation methods \cite{qi2017pointnet++,zhao2021point,hui2021pyramid}, the efficiency of our semantic feature-guided upsampling process is improved while ensuring that the semantic features of upsampling points are not easily smoothed out. 

%% file: result_discuss.tex
\section{Experiments}
\label{sec:Experiments}

In this section, we first present the implementation details of our method, including hardware configuration, training strategy, and  hyperparameter settings. Secondly, we present the performance of our network on two public segmentation datasets (ShapeNet \cite{wu20153d} and Toronto-3D \cite{tan2020toronto}, which are synthetic and real-scanned datasets respectively) and one remote sensing dataset \cite{zhao2021airborne}. We also compared our method with SOTA works in point cloud segmentation. Lastly, we conducted ablation studies to verify the effectiveness of each main component in our framework, which we present at the end of this section.

\subsection{Implementation Details}
We implemented DCTNet with PyTorch and trained it on an NVIDIA GeForce RTX 3090 GPU.
The network was trained with the Adam Optimizer, with a momentum of $0.9$ and weight decay of $0.0001$. The initial learning rate was set to $0.001$, with a cosine annealing schedule to adjust the learning rate at every epoch. The network was trained for $250$ epochs. The batch size was set as $16$.

\begin{figure*}[htbp]
  \centering
  \includegraphics[width=\linewidth]{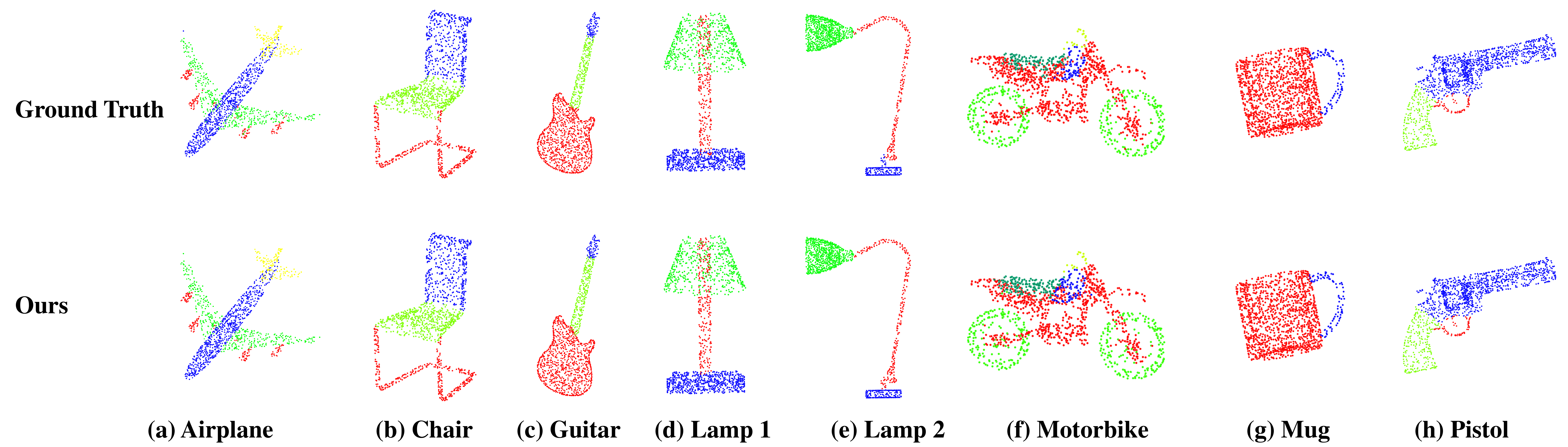}
  \caption{Part segmentation results from the ShapeNet dataset. As can be seen, our segmentation predictions are faithful to ground truth.
  \label{fig:shapenet}}
\end{figure*}

\subsection{Part Segmentation on ShapeNet Dataset}
\textbf{Datasets and Metrics.}
The ShapeNet dataset contains $16872$ synthetic models with 16 shape categories. They were split into $13998$ samples for training and $2874$ samples for testing, following Point Transformer\cite{zhao2021point}. This dataset has $50$ part labels, and each object has at least two parts. For a fair comparison, each input point cloud was downsampled to $2048$ points. 
For the evaluation metrics, we used the instance-wise mean Intersection over Union (instance-wise mIoU) and Frame Per Second to measure the accuracy and inference efficiency of algorithms respectively.

\textbf{Performance Comparison.}
We compared our DCTNet with SOTA segmentation methods. As shown in Fig. \ref{fig:eff} and Table \ref{tab:partseg_shapenet}\footnote{Missing entries are due to lack of source code for particular benchmarked models.}, DCTNet achieves the competitive instance-wise mIoU ($86.6\%$) with SOTA methods such as Point Transformer \cite{zhao2021point} and Stratified Transformer \cite{lai2022stratified}. 
As for the inference efficiency, compared with those algorithms (Point Transformer \cite{zhao2021point}, Stratified Transformer \cite{lai2022stratified}, etc.) that used FPS for downsampling, our method achieves a higher Frame Per Second (37). 
% This is because the proposed SuperiorPoint-based sampling and clustering are more concise than FPS, while facilitating similar semantic feature aggregating, instead of grouping semantic features according to 3D coordinates.
These results indicate our DCTNet greatly improves processing efficiency while maintaining accuracy. Several visual results of part segmentation are shown in Fig. \ref{fig:shapenet}. As can be seen, segmentation predictions from DCTNet are faithful to the ground truth. Moreover, Fig. \ref{fig:comparison} shows the airplane segmentation results of different methods. DCTNet achieves the best segmentation at the boundaries of adjacent parts (fuselage and wings). 
% This is because, for points located at the junction, the query ball grouping method used by PointNet++ \cite{qi2017pointnet++} and Point Transformer \cite{zhao2021point} tends to include points from different parts into the same group, disturbing the local feature aggregating. However, our SuperiorPoint-based clustering method in the LFA block focuses on grouping points with similar semantic information, which is able to ensure feature consistency within the same group.

% ！！！May add some comparion results with other methods in the fig, showing that our algorithm works better in border regions.

% show the SuperiorPoints

% show several SuperiorPoints in the fig, showing it contains similar semantic features. Compared with FPS, showing it will make mistakes in the border regions. OR this fig could be shown in the method to show the differece.

\begin{table}[htbp]\color{black}
 \caption{Part segmentation results on the ShapeNet dataset \label{tab:partseg_shapenet}
 }

 \centering
 \begin{tabular}{c|c|c}
  \hline
   {Methods} & Ins. mIoU ($\%$) & Frame Per Sec. \\
  \hline
  {PointNet}\cite{qi2017pointnet} & 83.7 & \textbf{200.0}    \\
  {PointNet++(SSG)}\cite{qi2017pointnet++} & 84.6 & 1.4   \\
  {PointNet++(MSG)}\cite{qi2017pointnet++} & 85.1 & 0.8   \\
  {PCNN}\cite{atzmon2018point} & 85.1  & -  \\
  {SpiderCNN}\cite{xu2018spidercnn} & 85.3 & 13.9   \\
  {SGPN}\cite{wang2018sgpn} & 85.8 & -  \\ 
  {PointCNN}\cite{li2018pointcnn} & 86.1 & -  \\
  {DGCNN}\cite{wang2019dynamic} & 85.2  & -  \\
  {PointConv}\cite{wu2019pointconv} & 85.7 & 15.9  \\
  {RS-CNN}\cite{liu2019relation} & 86.2  & 9.4 \\
  {KPConv}\cite{thomas2019kpconv} & 86.2  & 58.8  \\
  {InterpCNN}\cite{mao2019interpolated} & 86.3  & -  \\
  {DensePoint}\cite{liu2019densepoint} & 86.4 & -  \\
  {PAConv}\cite{xu2021paconv} & 86.1  & 50  \\
  {PointCloudTransformer}\cite{guo2021pct} & 86.4 & -  \\
  {PointTransformer}\cite{zhao2021point} & \textbf{86.6} & 7.8  \\
  {PointVoxelTransformer}\cite{zhang2022pvt} & 86.5 & 9.7  \\
  {StratifiedTransformer}\cite{lai2022stratified} & \textbf{86.6}  & 2.2  \\

  \hline
  {Ours}    & \textbf{86.6 }  & 37.0 \\

  \hline
 \end{tabular}
\end{table}

%#################################################
\subsection{Semantic Segmentation on Toronto-3D Dataset}

\textbf{Datasets and Metrics.}
Toronto-3D dataset was collected by a $32$-line LiDAR sensor in large-scale urban outdoor scenarios. It consists of above $78$ million points, covering approximately $1$ km of road. There are 8 categories included in the dataset: Road, Road marking, Natural, Building, Utility line, Pole, Car, and Fence. For fair comparison, we split the dataset into four subsets: $L001, L002, L003, L004$, where $L002$ was used for testing. Further, each subset was divided into a series of $5m \times 5m$ blocks, and each of them contained 2048 points after downsampling. To be consistent with the evaluation metrics used by SOTA methods, we used category-wise mIOU for performance evaluation. We also provide the IOU value for each category.

\begin{table*}[htbp]\color{black}
 \caption{Segmentation results ($\%$) on the Toronto-3D dataset\label{tab:semseg_toronto3d}
 }
 \centering
 \begin{tabular}{c|c|c|c|c|c|c|c|c|c}
  \hline
   {Methods}  & mIoU & Road & Road mrk.  & Natural & Building & Util.line & Pole & Car & Fence  \\
  \hline
  {PointNet++}\cite{qi2017pointnet++} & 56.55  & 91.44   & 7.59  & 89.80 &74.00 & 68.60 & 59.53 & 52.97 & 7.54     \\
  % {PointNet++ (MSG)}\cite{qi2017pointnet++} & 53.12  & 90.67   & 0.00  & 86.68 &75.78 & 56.20 & 60.89 & 44.51 & 10.19     \\
  {DGCNN}\cite{wang2019dynamic} & 49.60  & 90.63   & 0.44  & 81.25 &63.95 & 47.05 & 56.86 & 49.26 & 7.32     \\
  {KPFCNN}\cite{thomas2019kpconv} & 60.30  & 90.20   & 0.00  & 86.79 &86.83 & 81.08 & 73.06 & 42.85 & 21.57     \\
  {MS-PCNN}\cite{ma2019multi} & 58.01  & 91.22   & 3.50  & 90.48 &77.30 & 62.30 & 68.54 & 52.63 & 17.12     \\
  {TG-Net}\cite{li2019tgnet} & 58.34  & 91.39   & 10.62  & 91.02 &76.93 & 68.27 & 66.25 & 54.10 & 8.16     \\
  {MS-TGNet}\cite{tan2020toronto} & 60.96  & 90.89   & 18.78  & \textbf{92.18} &80.62 & 69.36 & 71.22 & 51.05 & 13.59     \\
  {Rim et al.}\cite{rim2021semantic} & 66.87	&\textbf{92.74}	&14.75	&88.66	&\textbf{93.52}	&81.03	&67.71	&39.65	&56.90     \\
  {diffConv}\cite{lin2021diffconv} & 76.73  & 83.31   & 51.06 &69.04 &79.55 & 80.48 & \textbf{84.41} & 76.19 & 89.83     \\
  {PCT}\cite{guo2021pct} & 79.32  & 79.77   & 59.51  & 75.78 &84.29 &77.78 &82.00 & 79.51 & 95.92     \\
  \hline
  {Ours}    & \textbf{81.84}  & 82.77  & \textbf{59.53} &85.51 &86.47 & \textbf{81.79} & 84.03 & \textbf{79.55} & \textbf{96.21}     \\

  \hline
 \end{tabular}
\end{table*}

\begin{table*}[htbp]\color{black}
 \caption{Segmentation results ($\%$) on the MS-LiDAR dataset \label{tab:semseg_remote}
 }
 \centering
 \begin{tabular}{c|c|c|c|c|c|c|c|c}
  \hline
   {Methods}  & Road & Building & Grass  & Tree & Soil & Powerline & OA & Average $F_{1}$   \\
  \hline
  {PointNet}\cite{qi2017pointnet} & 50.81  & 79.20   & 68.61  & 75.21 &12.73 & 22.56 & 83.36 & 51.52    \\
  {PointNet++}\cite{qi2017pointnet++} & 71.08  & 83.98   & 93.24  & 96.45 &30.24 & 57.28 & 90.43 & 72.05      \\
  {DGCNN}\cite{wang2019dynamic} & 70.42  & 90.25   & 93.62  & 97.93 &21.97 & 55.24 & 91.19 & 71.57     \\
  {RSCNN}\cite{liu2019relation} & 71.18  & 89.00   & 91.42  & 95.63 &26.43 & 70.03 & 92.44 & 73.90      \\
  {GACNet}\cite{wang2019graph} & 64.51  & 84.21   & 93.41  & 96.66 &22.77 & 33.83 & 87.59 & 67.65     \\
  {SE-PointNet++}\cite{jing2021multispectral} & 70.32  & 85.64   & 94.70  & 97.05 &37.02 & 70.35 & 93.01 & 75.84      \\
  {FR-GCNet}\cite{zhao2021airborne} & 82.63  & 90.81   & 95.33  & \textbf{98.77} &28.72 &\textbf{74.11} & 93.55 & 78.61     \\
  \hline
  {Ours}    & \textbf{87.64}  & \textbf{92.30}  & \textbf{98.81} &92.97 &\textbf{62.13} & 53.08 & \textbf{94.24} & \textbf{81.11}  \\

  \hline
 \end{tabular}
\end{table*}

%###########################################################

\textbf{Performance Comparison.}
We show the comparison results in Table \ref{tab:semseg_toronto3d}. DCTNet achieves the highest category-wise mIOU ($81.84\%$) compared to all benchmarked methods, surpassing SOTA methods such as diffconv \cite{lin2021diffconv}. Moreover, our method obtains the best IOUs in four (Road marking, Utility line, Car, and Fence) out of eight categories. These results demonstrate the superiority of our DCTNet in dealing with real-scanned data over previous SOTA methods.

\subsection{Airborne MS-LiDAR Dataset}

\textbf{Datasets and Metrics.}
Most recently, a large-scale airborne MultiSpectral LiDAR (MS-LiDAR) dataset was proposed in \cite{zhao2021airborne}. We tested DCTNet on this dataset to explore its performance in practical remote sensing applications. The MS-LiDAR dataset was captured by a Teledyne Optech Titan MS-LiDAR system \cite{zhao2021airborne}. In addition to three-dimensional coordinates, each point also has three channels with wavelengths of $1,550$ nm (MIR), $1,064$ nm (NIR), and $532$ nm (Green). There are six categories in the dataset: Road, Building, Grass, Tree, Soil, and Powerline. The dataset was divided into $13$ subsets, where subsets $1$-$10$ were taken as training data, while subsets $11$-$13$ were taken as testing data. For fair comparison, we took the same data pre-processing (data fusion, normalization, and training/testing sample generation) methods described in \cite{zhao2021airborne}. 
% Finally, there were 5300 samples for training, and 710 samples for testing (only Subset 11 was taken for testing, following \cite{zhao2021airborne}). 
Each sample contained $4096$ points with six channels. We used Overall Accuracy (OA), and average $F_{1}\mbox{-}score$ for performance evaluation, and provide the $F_{1}\mbox{-}score$ for each category.

\begin{figure}[htbp]
  \centering
  \includegraphics[width=\linewidth]{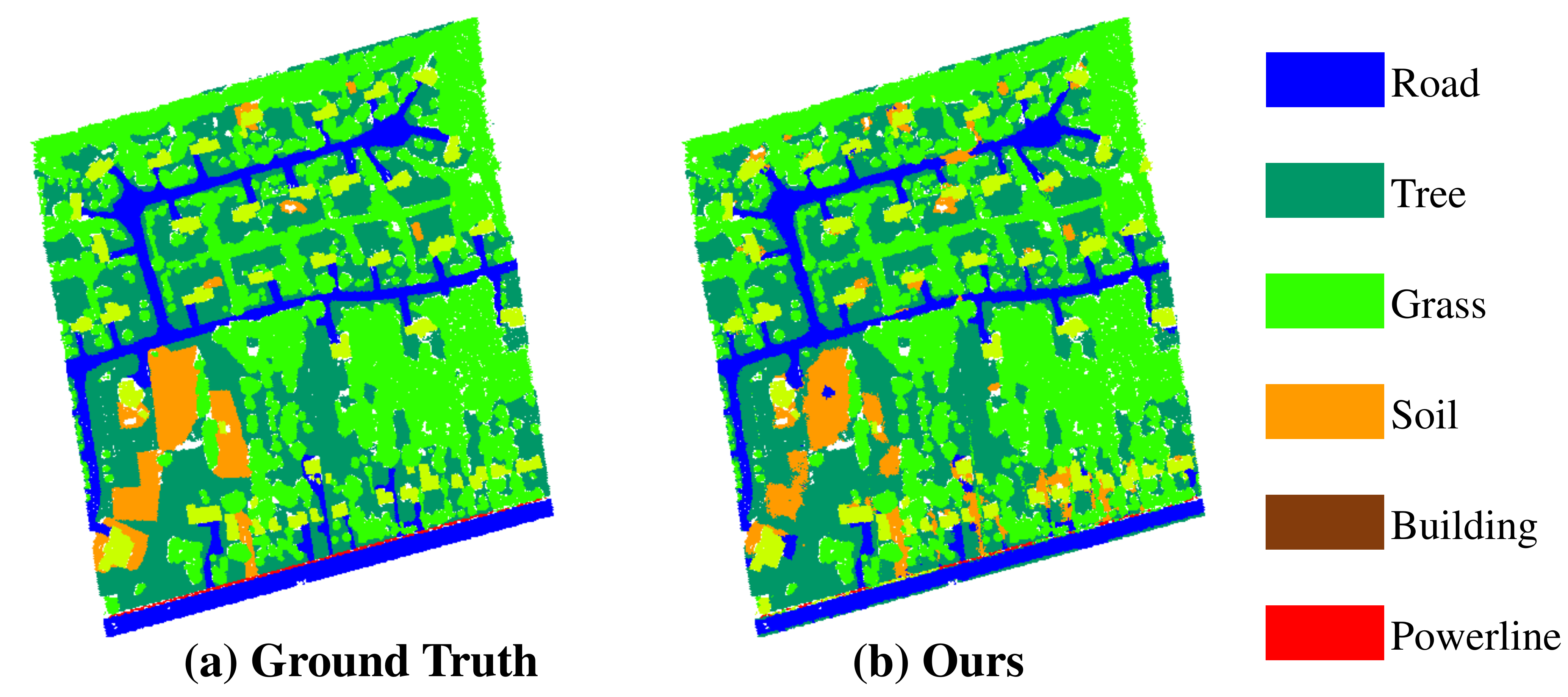}
  \caption{MS-LiDAR data segmentation results of our DCTNet.
  \label{fig:rs}}
\end{figure}

\textbf{Performance Comparison.}
The comparison results are shown in Table \ref{tab:semseg_remote}. Our DCTNet outperforms all benchmarked methods, achieving the best results in terms of both OA ($94.24\%$) and average $F_{1}\mbox{-}score$ ($81.11\%$). Moreover, our method also obtains the highest $F_{1}\mbox{-}score$s in four (Road, Building, Grass, and Soil) of six categories. Visual segmentation results are shown in Fig. \ref{fig:rs}. These results show that our DCTNet has an excellent performance in MS-LiDAR point cloud segmentation, exceeding that of previous methods.

%###########################################################

%#################################################

%#################################################

%#################################################
% \begin{table*}[htbp]\color{black}
%  \centering
%   \renewcommand{\arraystretch}{1.5} % Default value: 1
% \small
%  \caption{\textcolor{black}{Ablation study} 
%  }
%  \label{tab:ablation}
%  \begin{tabular}{c|c|c|c}
%   \hline
%     \multicolumn{2}{c|}{Ablation}  & Ins. mIOU & Frame Per Second \\
%  \hline
%   \multirow{3}{*}{Local feature aggregating} & SuperiorPoint-based sampling $\rightarrow$ FPS & 86.2  & 6.7     \\
%   \cline{2-4}
%       &SuperiorPoint-based clustering $\rightarrow$ $k$NN + MLP &86.2  & 19.5 \\
%    \cline{2-4}
%       &$-$ Cross-attention Transformer &85.7  & 42.1 \\
%    \hline
%    \multirow{2}{*}{Global feature learning} 
%       &$-$ CSA & 86.1  & 39.8 \\
%   \cline{2-4}
%      &$-$ PSA & 85.4   & \textbf{44.2} \\
%     \hline
%   \multirow{2}{*}{Upsampling} & Trilinear interpolation & 86.3 & 33.8     \\
%   \cline{2-4}
%      &Nearest neighbor interpolation & 86.4   & 35.3\\
%    \hline
%    \hline

%   \multicolumn{2}{c|}{3DPST}  & \textbf{86.6}   & 37.0  \\
%   \hline
%  \end{tabular}
% \end{table*}

\subsection{Ablation Study}
Ablation studies were conducted on ShapeNet dataset, to verify the effectiveness of main blocks in DCTNet. 

\textbf{Dynamic Clustering-based LFA Block.}
In the LFA block (Sec. \ref{subsec:LFA}) of DCTNet, we propose SDS and SDC methods for local feature aggregating, then use the cross-attention Transformer for feature enhancement. 

To evaluate their effectiveness, we first replaced the SDS method with FPS. As shown in Table \ref{tab:ablation} (Row 2), the DCTNet network with FPS obtains a lower instance-wise mIOU ($86.2\%$) than with the SDS method ($86.6\%$). 
% This is because FPS only focuses on the geometric properties of point clouds, ignoring their semantic properties. It weakens the attention of the network to some parts with significant semantic information. By contrast, SuperiorPoint-based sampling is a kind of feature-oriented sampling algorithm, where points with high feature saliency can be selected as the sampling points. 
Additionally, FPS is also very time-consuming, which reduces the inference efficiency of the network. As shown in Table \ref{tab:ablation} (Row 2), we can see the DCTNet with FPS obtains a much lower Frame Per Second ($6.7$) than the original one ($37.0$). 

Secondly, we used the "$k$-Nearest Neighborhood ($k$NN) $+$ MultiLayer Perceptron (MLP) $+$ Maxpooling" to replace the proposed SDC method, following the local feature extraction process in PointNet++ \cite{qi2017pointnet++}. Correspondingly, the point cloud upsampling method in the decoder was also replaced with the trilinear-interpolation upsampling. As shown in Table \ref{tab:ablation} (Row 3), after replacement, the accuracy of the network is reduced to $86.2\%$ from $86.6\%$, which highlights the importance of the SDC method. The main reason is that the $k$NN method achieves feature grouping only based on three-dimensional coordinates, ignoring local semantic homogeneity.
% Another reason is that the Maxpooling operation fails to consider all involved features for aggregation, resulting in inevitable information loss. By contrast, the proposed novel clustering method achieves feature grouping according to the feature similarities, ensuring the reliability of feature aggregation. Moreover, the learning-based weighted averaging algorithm is able to take all clustering points into account, avoiding the information loss caused by the Maxpooling. 
In terms of inference efficiency, Table \ref{tab:ablation} (Row 3) shows that the Frame Per Second ($19.5$) of the network after replacement is also lower than the original one ($37.0$). These results demonstrate that the proposed SDC method is able to improve local feature aggregation.

Finally, we removed the cross-attention Transformer in the LFA block. As shown in Table \ref{tab:ablation} (Row 4), without the cross-attention Transformer, the category-wise mIOU is reduced from $86.6\%$ to $85.7\%$.%, indicating the effectiveness of this component.

\textbf{Dual-attention Transformer-based GFL Block.}
Dual-attention Transformers (Sec. \ref{subsec:GFL}) have been proven effective by previous works \cite{han2022dual,lu20223dpct}. We conducted ablation studies to verify that dual-attention Transformers outperformed vanilla Transformers with only point-wise or channel-wise self-attention mechanisms. As shown in Table \ref{tab:ablation} (Row 5), when the channel-wise self-attention is removed, the accuracy of DCTNet drops from $86.6\%$ to $86.1\%$. Similarly, when the point-wise self-attention is removed (Table \ref{tab:ablation}, Row 6), there is a similar drop in terms of accuracy (from $86.6\%$ to $85.4\%$). %These results demonstrate that, in our network, the dual-attention Transformer works better than vanilla Transformers.

\textbf{Point Cloud Upsampling Block.}
We compared the proposed semantic feature-guided upsampling (Sec. \ref{subsec:upsampling}) in the decoder with several commonly used upsampling methods, such as trilinear interpolation and nearest neighbor interpolation. The results are shown in Table \ref{tab:ablation} (Row 7, 8). According to the results, our semantic feature-guided upsampling method outperforms the aforementioned two interpolation methods in terms of both the category-wise mIOU and Frame Per Second. This is because our upsampling method assigns the features of the sampling points to the corresponding cluster points, according to the relationship stored in the encoder. This ensures that the semantic features of upsampled points are not easily smoothed out and slightly improves upsampling efficiency. %Additionally, the efficiency of the upsampling process is also slightly improved.

\begin{table*}[htbp]\color{black}
 \centering
 \caption{\textcolor{black}{Results of ablation studies} 
 }
 \label{tab:ablation}

 \begin{tabular}{c|c|c|c}
  \hline
    \multicolumn{2}{c|}{Ablation}  & Ins. mIOU ($\%$) & Frame Per Sec. \\
 \hline
  \multirow{3}{*}{LFA} & FPS & 86.2  & 6.7     \\
  \cline{2-4}
      & $k$NN + MLP &86.2  & 19.5 \\
   \cline{2-4}
      &$-$ Cross-attention Transformer &85.7  & 42.1 \\
   \hline
   \multirow{2}{*}{GFL} 
      &$-$ CSA & 86.1  & 39.8 \\
  \cline{2-4}
     &$-$ PSA & 85.4   & \textbf{44.2} \\
    \hline
  \multirow{2}{*}{Upsampling} & Trilinear interpolation & 86.3 & 33.8     \\
  \cline{2-4}
     &Nearest neighbor interpolation & 86.4   & 35.3\\
   \hline
   \hline

  \multicolumn{2}{c|}{DCTNet}  & \textbf{86.6}   & 37.0  \\
  \hline
 \end{tabular}
\end{table*}

%% file: conclusion.tex
\section{Conclusion}
\label{sec:conclusion}
In this paper, we propose DCTNet, a novel Transformer-based 3D point cloud processing framework for semantic segmentation. DCTNet has a hierarchical encoder-decoder structure. %By design, the encoder conducts both local and global feature learning.
For local feature learning, we propose the new Semantic feature-based Dynamic Sampling and Clustering algorithms, acronymed as SDS and SDC respectively. Compared with prevalent sampling and grouping methods, our SDS and SDC are more suitable for semantic information learning, while also facilitating the point cloud upsampling process. For global feature learning, we utilize dual-attention Transformer blocks which excel at modelling long-range dependencies.
% , due to their remarkable ability of long-range context dependency modeling
Our decoder is symmetric to the encoder, but contains our newly designed semantic feature-guided upsampling method which improves efficiency and ensure that the semantic features of upsampling points are not easily smoothed out. Extensive experiments on the ShapeNet \cite{wu20153d} and Toronto-3D datasets \cite{tan2020toronto} demonstrate that DCTNet outperforms previous methods. For example, the inference speed of DCTNet is 3.8-16.8$\times$ faster than existing SOTA models on the ShapeNet dataset, while achieving the instance-wise mIoU top score of $86.6\%$.
We also tested DCTNet to a remote sensing dataset \cite{zhao2021airborne}, demonstrating its excellent performance in segmenting MS-LiDAR data. These results show that DCTNet has achieved State-of-the-Art status in point cloud segmentation.